\documentclass{doublecol}
\usepackage{amssymb}
\usepackage{amsmath}
\usepackage{latexsym}
\usepackage{theorem}
\usepackage{graphicx}
\usepackage{subfigure}
\usepackage{graphics}
\usepackage{epsfig}
\usepackage{listings}

\usepackage{amsbsy,natbib}
\usepackage{amsfonts,bm}

\RequirePackage{graphics,epsf}

\lstdefinelanguage{Pascal}{morekeywords={repeat,false,self,super,nil}}

\newtheorem{definition}{Def\-inition}
\renewcommand{\lstlistingname}{Algorithm}

\begin{document}


%
%
%
%
%
%
%
%
%
%
%
%
\title{Review and Evaluation of Feature Selection Algorithms in Synthetic Problems}

\authorA{L.A. Belanche*}

\affA{Dept. de  Llenguatges i Sistemes Inform\`atics, \\
Universitat Polit\`ecnica de Catalunya, Barcelona, Spain \\ E-mail: belanche@lsi.upc.edu \\
*Corresponding author}

\authorB{F.F. Gonz\'alez}

\affB{Dept. de  Llenguatges i Sistemes Inform\`atics, \\
Universitat Polit\`ecnica de Catalunya, Barcelona, Spain \\ E-mail: fgonzalez@lsi.upc.edu}

\begin{abstract}
  The main purpose of Feature Subset Selection is to find a reduced
  subset of attributes from a data set described by a feature set. The
  task of a \textit{feature selection algorithm} (FSA) is to provide
  with a computational solution motivated by a certain definition of
  \emph{relevance} or by a reliable evaluation measure. In this paper
  several fundamental algorithms are studied to assess their
  performance in a controlled experimental scenario. A measure to
  evaluate FSAs is devised that computes the degree of matching
  between the output given by a FSA and the known optimal solutions.
  An extensive experimental study on synthetic problems is carried out
  to assess the behaviour of the algorithms in terms of solution
  accuracy and size as a function of the relevance, irrelevance,
  redundancy and size of the data samples. The controlled experimental
  conditions facilitate the derivation of better-supported and
  meaningful conclusions.
\end{abstract}

\KEY{Feature Selection Algorithms; Empirical Evaluations; Attribute relevance and redundancy.}


\maketitle

\section{INTRODUCTION}
\label{section:Introduction}

The \textit{feature selection} problem is ubiquitous in an inductive
machine learning or data mining setting and its importance is beyond
doubt.  The main benefit of a correct selection is the improvement of
the inductive learner, either in terms of learning speed,
generalization capacity or simplicity of the induced model. On the
other hand, there are the \textit{scientific} benefits associated with
a smaller number of features: a reduced measurement cost and hopefully
a better understanding of the domain.  A {\em feature selection
  algorithm} (FSA) is a computational solution that should be guided
by a certain definition of \emph{subset relevance}, although in many
cases this definition is implicit or followed in a loose sense.  This
is so because, from the inductive learning perspective, the relevance
of a feature may have several definitions depending on the precise
objective that is looked for \citep{Caruana:1994}. Thus the need
arises to count on common sense criteria that enables to adequately
decide which algorithm to use (or {\em not} to use) in certain
situations.

This work reviews the merits of several fundamental feature subset
selection algorithms in the literature and assesses their performance
in an artificial controlled experimental scenario. A \textit{scoring
  measure} computes the degree of matching between the output given by
the algorithm and the known optimal solution. This measure ranks the
algorithms by taking into account the amount of relevance, irrelevance
and redundancy on synthetic data sets of discrete features. Sample
size effects are also studied.

The results illustrate the strong dependence on the particular
conditions of the algorithm used, as well as on the amount of
irrelevance and redundancy in the data set description, relative to
the total number of features. This should prevent the use of a single
algorithm specially when there is poor knowledge available about the
structure of the solution. More importantly, it points in the
direction of using principled combinations of algorithms for a more
reliable assessment of feature subset performance.

The paper is organized as follows: we begin in Section
\ref{section:relatedwork} reviewing relevant related work. In section
\ref{section:DefinicionFS} we set a precise definition of the
feature selection problem and briefly survey the main categorization
of feature selection algorithms. We then provide an algorithmic
description and comment on several of the most widespread algorithms
in section \ref{section:algoritmos}. The methodology and tools used
for the empirical evaluation are covered in section
\ref{section:experimental-designer}. The experimental study, its
results and a general advice to the data mining practitioner are
developed in section \ref{section:exp-eval}. The paper ends with the
conclusions and prospects for future work.

\section{MOTIVATION AND RELATED WORK}
\label{section:relatedwork} Previous experimental work on feature
selection algorithms for comparative purposes include \cite{Aha:1995},
\cite{Doak:1992}, \cite{Jain:1997}, \cite{Kudo:1997} and
\cite{LiuHa:1998}. Some of these studies use artificially generated
data sets, like the widespread \textsl{Parity}, \textsl{Led} or
\textsl{Monks} problems \citep{Thrun:1991}.  Demonstrating improvement
on synthetic data sets can be more convincing that doing so in typical
scenarios where the true solution is completely unknown. However,
there is a consistent lack of systematical experimental work using a
common benchmark suite and equal experimental conditions.  This
hinders a wider exploitation of the power inherent in fully controlled
experimental environments: the knowledge of the (set of) optimal
solution(s), the possibility of injecting a desired amount of
relevance, irrelevance and redundancy and the unlimited availability
of data.

Another important issue is the way FSA performance is assessed. This
is normally done by handing over the solution encountered by the FSA
to a specific inducer (during of after the feature selection process
takes place). Leaving aside the dependence on the particular inducer
chosen, there is a much more critical aspect, namely, the relation
between the performance as reported by the inducer and the true merits
of the subset being evaluated. In this sense, it is our hypothesis
that FSAs are very affected by finite sample sizes, which distort
reliable assessments of subset relevance, even in the presence of a
very sophisticated search algorithm \citep{Reunanen:2003}.  Therefore,
sample size should also be a matter of study in a through experimental
comparison. This problem is aggravated when using filter measures,
since in this case the relation to true generalization ability (as
expressed by the {\em Bayes error}) can be very loose
\citep{Ben-Bassat:1982}.

A further problem with traditional benchmarking data sets is the
implicit assumption that the used data sets are actually {\em
  amenable} to feature selection. By this it is meant that performance
benefits clearly from a good selection process (and less clearly or
even worsens with a bad one).  This criterion is not commonly found in
similar experimental work. In summary, the rationale for using
exclusively synthetic data sets is twofold:

\begin{enumerate}
\item Controlled studies can be developed by systematically varying
  chosen experimental conditions, thus facilitating the derivation of
  more meaningful conclusions.

\item Synthetic data sets allow full control of the experimental
  conditions, in terms of amount of relevance, irrelevance and
  redundancy, as well as sample size and problem difficulty. An added
  advantage is the knowledge of the set of optimal solutions, in which
  case the degree of closeness to any of these solutions can thus be
  assessed in a confident and automated way.
\end{enumerate}

The procedure followed in this work consists in generating sample data
sets from synthetic functions of a number of discrete relevant
features. These sample data sets are then corrupted with irrelevant
and/or redundant features and handed over to different FSAs to
obtained a hypothesis. A {\em scoring measure} is used in order to
compute the degree of matching between this hypothesis and the known
optimal solution. The score takes into account the amount of
relevance, irrelevance and redundancy in each suboptimal solution as
yielded by an algorithm.

The main criticism associated with the use of artificial data is the
likelihood that such a problem be found in real-world scenarios. In
our opinion this issue is more than compensated by the mentioned
advantages. A FSA that is not able to work properly in simple
experimental conditions (like those developed in this work) is in
strong suspect of being inadequate in general.

\section{THE FEATURE SELECTION PROBLEM}
\label{section:DefinicionFS}

Let $X$ be the original set of features, with cardinality $|X| = n$.
The \emph{continuous} feature selection problem (also called Feature
Weighing) refers to the assignment of weights $w_i$ to each feature
$x_i \in X$ in such a way that the order corresponding to its
theoretical relevance is preserved. The \emph{binary} feature
selection problem (also called Feature Subset Selection) refers to the
choice of a subset of features that jointly maximize a certain measure
related to subset relevance. This can be carried out directly, as many
FSAs do \citep{Almuallim:1991,Caruana:1994,Hall:1999}, or setting a
\emph{cut-point} in the output of the continuous problem solution.
Although both types can be seen in an unified way (the latter case
corresponds to the assignment of weights in $\{0,1\}$), these are
quite different problems that reflect different design objectives. In
the continuous case, one is interested in keeping {\em all} the
features but in using them {\em differentially} in the learning
process. On the contrary, in the binary case one is interested in
keeping just a {\em subset} of the features and (most likely) using
them {\em equally} in the learning process.

A common instance of the feature selection problem can be formally
stated as follows.  Let $J$ be a performance evaluation measure to be
optimized (say to maximize) defined as $J:{\cal P}(X) \rightarrow
\mathbb{R}^+\cup \{0\}$. This function accounts for a general
evaluation measure, that may or may not be inspired in a precise and
previous definition of relevance.  Let $c(x) \geq 0$ represent the
{\em cost} of variable $x$ (measurement cost, needed technical skill,
etc) and call $c(X') = \sum\limits_{x \in X'} c(x)$, for $X' \in {\cal
  P}(X)$. Let $C_X = c(X)$ be the cost of the whole feature set. It is
assumed here that $c$ is additive, that is, $c(X' \cup X'') = c(X') +
c(X'')$ (together with non-negativeness, this implies that $c$ is
monotone).

\begin{definition}[Feature Subset Selection] \label{def-fs} The selection of an optimal feature
subset (``Feature Subset Selection") is either of two scenarios:

\begin{itemize}

 \item[(1)] Fix $C_0 \leq C_X$. Find the $X' \in {\cal P}(X)$ of maximum
$J(X')$ among those that fulfill $c(X') \leq C_0$.

 \item[(2)] Fix $J_0 \geq 0$. Find the $X' \in {\cal P}(X)$ of minimum
$c(X')$ among those that fulfill $J(X') \geq J_0$.

\end{itemize}
\end{definition}

If the costs are unknown, a meaningful choice is obtained by setting
$c(x) = 1$ for all $x \in X$.  Then $c(X')=|X'|$ and $c$ can be
interpreted as the {\em complexity} of the solution. In this case, (1)
amounts to finding the subset with highest $J$ among those having a
maximum pre-specified size.  In scenario (2), it amounts to finding
the smallest subset among those having a minimum pre-specified
performance (as measured by $J$). Only with these restrictions, an
optimal subset of features need not exist; if it does, is not
necessarily unique.  In scenario (1), a solution always exists by
defining $c(\emptyset) = 0$ and $J(\emptyset)$ to be the value of $J$
with no features.  In case (2), if there is no solution, an adequate
policy may be to set $J_0 = \epsilon J(X), \epsilon>0$, and
progressively lower the value of $\epsilon$.  If there is more than
one solution (of equal performance and cost, by definition) one is
usually interested in them all. We shall speak of a FSA of {\em Type
  1} (resp. {\em Type 2}) when it has been designed to solve the first
(resp. second) scenario in Def.  \ref{def-fs}. If the FSA can be used
in both scenarios, we shall speak of a {\em general-type} algorithm.
In addition, if one has no control whatsoever, we shall speak of a
{\em free-type} algorithm.

We shall use the notation $S(X')$ to indicate the
subsample of $S$ described by the features in $X' \subseteq X$ only.

\section{FEATURE SUBSET SELECTION ALGORITHMS}
\label{section:algoritmos}

The relationship between a FSA and the inductive learning method used
to infer a model can take three main forms: \emph{filter},
\emph{wrapper} or \emph{embedded}, which we call the {\em mode}:

\medskip\noindent \textbf{Embedded Mode:} The inducer has its own FSA
(either explicit or implicit).  The methods to induce logical
conjunctions \citep{Vere:1975,Winston:1975}, decision trees or
artificial neural networks are examples of this embedding.

\medskip\noindent \textbf{Filter Mode:} If feature selection takes
place before the induction step, the former can be seen as a filter
(of non-useful features).  In a general sense it can be seen as a
particular case of the embedded mode in which feature selection is
used as a pre-processing. The filter mode is then independent of the
inducer that evaluates the model after the feature selection process.

\medskip\noindent \textbf{Wrapper Mode:} Here the relationship is
taken the other way around: the FSA uses the learning algorithm as a
subroutine \citep{John:1994}. The argument in favor of this mode is to
{\em equal the bias} of the FSA and the inducer that will be used
later to assess the goodness of the model. A main disadvantage is the
computational burden that comes from calling the inducer to evaluate
each and every subset of considered features.

In what follows several of the currently most widespread FSAs in
machine learning are described and briefly commented on.
General-purpose search algorithms, as genetic algorithms, are excluded
from the review. None of the algorithms allow the specification of
costs in the features. Most of them can work in filter or wrapper
mode. The {\em feature weighing} algorithm \textsc{Relief} has been
included, both in the review and in the experimental comparison, as a
complement.  This is so because it can also be used to select a subset
of features, although a way of getting a subset out of the weights has
to be devised. In the following we assume again that the evaluation
measure $J$ is to be maximized.

\subsection{Algorithm LVF}
\label{section:lvf}

\textsc{Lvf} (Las Vegas Filter) \citep{LVF-LVI:1998} is a type 2
algorithm that repeatedly generates random subsets and computes the
\textit{consistency} of the sample: an \emph{inconsistency} in $X'$
and $S$ is defined as two instances in $S$ that are equal when
considering only the features in $X'$ and that belong to different
classes. The aim is to find the \emph{minimum} subset of features
leading to zero inconsistencies. The {\em inconsistency count} of an
instance $A \in S$ is defined as:
\begin{equation}
IC_{X'}(A) = X'(A) - \max_k X'_k(A)
\end{equation}
 \noindent
 where $X'(A)$ is the number of instances in $S$ equal to
$A$ using only the features in $X'$ and $X'_k(A)$ is the number of
instances in $S$ of class $k$ equal to $A$ using only the features
in $X'$ \citep{LiuHa:1998}. The {\em inconsistency rate} of a feature subset in a
sample $S$ is then:
\begin{equation}
IR (X') = \frac{\sum_{A \in S} IC_{X'}(A) }{|S|}
\end{equation} \noindent This is a monotonic measure, in the sense
$$ X_1 \subset X_2 \Rightarrow IR (X_1) \geq IR (X_2) $$
The evaluation measure is then $J(X') = \frac{1}{IR (X') +
1} \in (0,1]$ that can be evaluated in $O(|S|)$
time using a hash table.

\textsc{Lvf} is described as Algorithm \ref{fig:algoritmo-lvf}. It has
been found to be particularly efficient for data sets having redundant
features \citep{DashL:1997}. Arguably its main advantage may be that
it quickly reduces the number of features in the initial stages with
certain confidence \citep{Dash:1998, Dash:2000}; however, many poor
solution subsets are analyzed, wasting computing resources.

\begin{lstlisting}[float=htb!,
 caption={\textsc{Lvf} (Las Vegas Filter).},
 label=fig:algoritmo-lvf, keywordstyle=\bfseries,
 texcl,frame=lines,captionpos=b,
 basicstyle=\footnotesize, mathescape]

 Input:
   $max$ - the maximum number of iterations
   $J$ - evaluation measure
   $S(X)$ - a sample $S$ described by $X$, $|X|=n$
 Output:
   $L$ - all equivalent solutions found

 $L$ := []      // L stores equally good sets
 $Best$ := $X$        // Initialize best solution
 $J_0$ := $J(S(X))$     // minimum allowed value of J
 repeat $max$ times
    $X'$ := Random_SubSet($Best$)
    if $J(S(X')) \geq J_0$   then
        if $|X'| < |Best|$ then
            $Best$ := $X'$
            $L$ := [$X'$]          // L is reinitialized
        else
           if $|X'| = |Best|$ then
              $L$ := append($L,X'$)
           end
        end
    end
 end
\end{lstlisting}
\subsection{Algorithm LVI}
\label{section:lvi}

\textsc{Lvi} (Las Vegas Incremental) is also a type 2 algorithm and an
evolution of \textsc{Lvi}.  It is based on the grounds that it is not
necessary to use the whole sample $S$ in order to evaluate the measure
$J$, which for this algorithm is again {\em consistency}
\citep{LiuHa:1998}.  The algorithm is described as Algorithm
\ref{fig:algoritmo-lvi}. It departs from a portion $S_0$ of $S$; if
\textsc{Lvf} finds a sufficiently good solution in $S_0$ then
\textsc{Lvi} halts. Otherwise the set of instances in $S \setminus S_0$
making $S_1$ inconsistent is added to $S_0$, this new portion is
handed over to \textsc{Lvf} and the process is iterated. Intuitively,
the portion cannot be too small or too big. If it is too small, after
the first iteration many inconsistencies will be found and added to
the current portion, which will hence be very similar to $S$. If it
is too big, the computational savings will be modest.  The authors
suggest $p = 10\%$ or a value proportional to the number of features.
In \cite{Liu:98} it is reported experimentally that \textsc{Lvi}
adequately chooses relevant features, but may fail for noisy
data sets, in which case the algorithm it is shown to consider
irrelevant features. Probably \textsc{Lvi} is more sensible to noise
than \textsc{Lvf} in cases of small sample sizes.

\begin{lstlisting}[float=htb!,
 caption={\textsc{Lvi} (Las Vegas Incremental).},
 label=fig:algoritmo-lvi,
 texcl,frame=lines,captionpos=b,
 basicstyle=\footnotesize,  mathescape]

 Input:
   $max$ - the maximum number of iterations
   $J$ - evaluation measure
   $S(X)$ - a sample $S$ described by $X, |X|=n$
   $p$ - initial percentage
 Output:
   $X'$ - solution found

 $S_0$ := portion($S,p$) // Initial portion
 $S_1$ := $S \setminus S_0$         
 $J_0$ := $J(S(X))$              // Minimum allowed value of J
 repeat forever
    $X'$ := LVF ($max,J,S_0(X)$)
    if $J(S_1(X')) \geq J_0$ then stop
    else
        $C$ := {$x \in S_1(X')$ making $S_1(X')$ inconsistent}
        $S_0$ := $S_0\ \cup C$ 
        $S_1$ := $S_1 \setminus C$
    end
 end
\end{lstlisting}

\subsection{Algorithm RELIEF}
\label{section:relief}

\textsc{Relief} \citep{Kira:1992} is a general-type algorithm that
works exclusively in filter mode. The algorithm randomly chooses an
instance $I \in S$ and finds its \emph{near hit} and its \emph{near
  miss}.  The former is the closest instance to $I$ among all the
instances in the same class of $I$. The latter is the closest instance
to $I$ among all the instances in a different class. The underlying
idea is that a feature is more relevant to $I$ the more it separates
$I$ and its near miss, and the least it separates $I$ and its near
hit. The result is a weighed version of the original feature set. The
algorithm for two classes is described as Algorithm \ref{fig:algoritmo-relief}.

\begin{lstlisting}[float=htb!,
 caption={\textsc{Relief}.},
 label=fig:algoritmo-relief,
 texcl,frame=lines,captionpos=b,
 basicstyle=\footnotesize, mathescape]

 Input:
   $p$ - sampling percentage
   $d$ - distance measure
   $S(X)$ - a sample $S$ described by $X, |X|=n$
 Output:
   $w$ - array of feature weights

 let $m$ := $p|S|$
 initialize array $w$[] to zero
 do m times
    $I$ := Random_Instance $(S)$
    $I_{nh}$ := Near-Hit $(I, S)$
    $I_{nm}$ := Near-Miss $(I, S)$
    for each $i \in [1..n]$ do
      $w[i]$ := $w[i]+d_i(I,I_{nm})/m - d_i(I,I_{nh})/m$
    end
 end
\end{lstlisting}

When costs are just sizes, the algorithm can be used to simulate a
type 1 scenario by iteratively checking the sequence of the first
$C_0$ nested subsets in the order given by decreasing weights, calling
the $J$ measure, and returning that subset with the highest value of
$J$. To simulate a type 2 scenario, the same sequence is checked
looking for the first element in the sequence that yields a value of
$J$ not less than the chosen $J_0$. The more important advantage of
\textsc{Relief} is the rapid assessment of irrelevant features with a
principled approach; however it does not make a good discrimination
among redundant features. The algorithm has been found to choose
correlated features instead of relevant features \citep{DashL:1997},
and therefore the optimal subset can be far from assured
\citep{Kira:1992}.  Some variants have been proposed to account for
several classes \citep{Kononenko:1994}, where the $k$ more similar
instances are selected and their averages computed.

\subsection{Algorithms SFG/SBG}
\label{section:sfg}

These two are classical general-type algorithms that may work in
filter or wrapper mode.  \textsc{Sfg} (Sequential Forward Generation)
iteratively adds features to an initial subset, trying to improve a
measure $J$, always taking into account those features already
selected. Consequently, an ordered list can also be obtained.
\textsc{Sbg} (Sequential Backward Generation) is the backward
counterpart. They are jointly described as Algorithm
\ref{fig:algoritmo-sfg-sbg}. When the number of features is small,
\cite{Doak:1992} reported that \textsc{Sbg} tends to show better
performance than \textsc{Sfg}, most likely because \textsc{Sbg}
evaluates the contribution of all features from the onset. In
addition, \cite{Aha:1995} points out that \textsc{Sfg} is preferable
when the number of relevant features is (known to be) small; otherwise
\textsc{Sbg} should be used. Interestingly, it was also reported that
\textsc{Sbg} did not \textit{always} have better performance than
\textsc{Sfg}, contrary to the conclusions in \cite{Doak:1992}.
Besides, \textsc{Sfg} is faster in practice.  The algorithms
\textsc{W-Sfg} and \textsc{W-Sbg} (W for \emph{wrapper}) use the
accuracy of an inducer as evaluation measure.

\begin{lstlisting}[float=htb!,
 caption={\textsc{Sbg/Sfg} (Sequential Backward/Forward
 Generation).},
 label=fig:algoritmo-sfg-sbg,
 texcl,frame=lines,captionpos=b,
 basicstyle=\footnotesize, mathescape]

 Input:
   $S(X)$ - a sample $S$ described by $X, |X|=n$
   $J$ - evaluation measure
 Output:
   $X'$ - solution found

 $X' := \emptyset$ /* Forward */ or $X' := X$ /* Backward */
 repeat
  $x'\,:=\,argmax\{J(S(X' \cup \{x\}))\,|\,x \in X \setminus X'\}\,$ /* Forward */
  $x'\,:=\,argmax\{J(S(X' \setminus \{x\}))\,|\,x \in X'\}\,$       /* Backward */
  $X' := X' \cup \{x'\}$        /* Forward */
  $X' := X' \setminus \{x'\}$   /* Backward */
 until no improvement in $J$ in last $j$ steps
       or $X'=X$ /*Forward*/ or $X'=\emptyset$ /*Backward*/
\end{lstlisting}

\subsection{Algorithms SFFG/SFBG}
\label{section:sffs}

These are free-type algorithms that may work in filter or wrapper
mode. \textsc{Sffg} (Sequential Floating Forward Generation)
\citep{Pudil:1994} is an exponential cost algorithm that operates in a
sequential fashion, performing a forward step followed by a variable
(and possibly null) number of backward ones. In essence, a feature is
first unconditionally added and then features are removed as long as
the generated subsets are the best among their respective size. The
algorithm (described in Algorithm 5 as a flow-chart) is so-called
because it has the characteristic of \emph{floating} around a
potentially good solution of the specified size. The backward
counterpart \textsc{Sfbg} performs a backward step followed by zero or
more forward steps. These two algorithms have been found to be very
effective in some situations \citep{Jain:1997}, and are among the most
popular nowadays. Their main drawbacks are the computational cost,
that may be unaffordable when the number of features nears the hundred
\citep{Bins:2001} and the need to fix the size of the final desired
subset.

\setcounter{figure}{4}
\renewcommand{\figurename}{Algorithm}

\begin{figure}
\begin{flushleft}
\rule{\linewidth}{0,4pt}
\end{flushleft}

\footnotesize{
Input:

   $\qquad S(X)$ - a sample $S$ described by $X, |X|=n$

   $\qquad J$ - evaluation measure

   $\qquad d$ - desired size of the solution

   $\qquad \Delta$ - maximum deviation allowed with respect to $d$

 Output:

   $\qquad $ solution of size $d \pm \Delta$
}

\normalsize
\begin{center}
    \epsfig{file=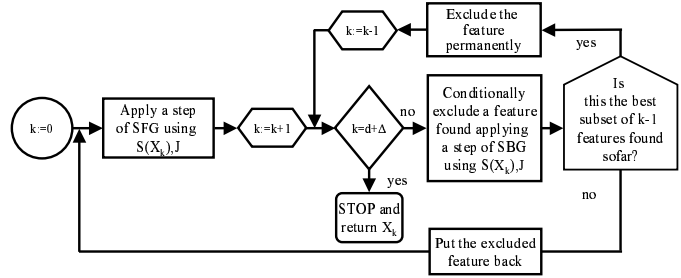, width=9.0cm}
\end{center}
\caption{\textsc{Sffg} (Sequential Floating Forward Generation). The
set $X_k$ denotes the current solution (of size $k$); $S(X_k)$ is the
sample described by the features in $X_k$ only.}
\begin{flushleft}
\rule{\linewidth}{0,4pt}
\end{flushleft}

\end{figure}



\subsection{Algorithm QBB}
\label{section:qbb}

The \textsc{Qbb} (Quick Branch and Bound) algorithm \citep{Dash:1998}
(described as Algorithm \ref{fig:algoritmo-qbb}) is a type 1
algorithm. Actually it is a hybrid one, composed of \textsc{Lvf} and
\textsc{Abb}. The origin of \textsc{Abb} is in Branch \& Bound
\citep{Narendra:1977}, an optimal search algorithm.  Given a threshold
$\beta$ (specified by the user), the search stops at each node the
evaluation of which is lower than $\beta$, so that efferent branches
are pruned. \textsc{Abb} (Automatic Branch \& Bound) \citep{Liu:1998}
is a variant having its bound as the inconsistency rate of the data
when the full set of features is used (Algorithm
\ref{fig:algoritmo-abb}).  The basic idea of \textsc{Qbb} consists in
using \textsc{Lvf} to find good starting points for \textsc{Abb}.  It
is expected that \textsc{Abb} can explore the remaining search space
efficiently. The authors reported that \textsc{Qbb} is, in general,
more efficient than \textsc{Lvf} or \textsc{Abb} in terms of average
cost of execution and selected relevant features.

\setcounter{lstlisting}{5}
\begin{lstlisting}[float=htb!,
 caption={\textsc{Abb} (Automatic Branch and Bound).},
 label=fig:algoritmo-abb,
 texcl,frame=lines,captionpos=b,
 basicstyle=\footnotesize, mathescape]

 Input:
   $S(X)$ - a sample $S$ described by $X, |X|=n$
   $J$ - evaluation measure (monotonic)
 Output:
   $L$ - all equivalent solutions found

 procedure ABB ($S(X)$: sample; var $L'$: list of set)
   for each $x$ in $X$ do
     enqueue($Q,X \setminus \{x\}$) // remove a feature at
                          a time
   end
   while not empty($Q$) do
     $X' :=$ dequeue($Q$)
     // $X'$ is legitimate if it is not a subset of
        a pruned state
     if legitimate($X'$) and $J(S(X')) \geq J_0$ then
        $L' :=$ append($L', X'$)
        ABB($S(X'), L'$)
     end
   end
 end

 begin
    $Q := \emptyset$            // Queue of pending states
    $L' := [X]$            // List of solutions
    $J_0 := J(S(X))$   // Minimum allowed value of $J$
    ABB $\ (S(X), L')$ // Initial call to ABB
    $k :=$ smallest size of a subset in $L'$
    $L :=$ set of elements of $L'$ of size $k$
 end
 \end{lstlisting}

\begin{lstlisting}[float=htb!,
 caption={\textsc{Qbb} (Quick Branch and Bound).},
 label=fig:algoritmo-qbb,
 texcl,frame=lines,captionpos=b,
 basicstyle=\footnotesize, mathescape]

 Input:
   $max$ - the maximum number of iterations
   $J$ - monotonic evaluation measure
   $S(X)$ - a sample $S$ described by $X, |X|=n$
 Output:
   $L$ - all equivalent solutions found

 $L\_ABB$ := []
 $L\_LVF$ := LVF ($max,J,S(X)$)
 for each $X' \in L\_LVF$ do
    $L\_ABB$ := concat($L\_ABB,ABB(S(X'),J)$)
 end
 $k$ := smallest size of a subset in $L\_ABB$
 $L$ := set of elements of $L\_ABB$ of size $k$
\end{lstlisting}

\section{EMPIRICAL EVALUATION OF FSAs}
\label{section:experimental-designer}

The main question arising in a feature selection experimental design
is: what are the aspects that we would like to evaluate of a FSA
solution in a given data set? Certainly a {\em good} algorithm is one
that maintains a well-balanced trade-off between small-sized and
competitive solutions. To assess these two issues at the same time is
a difficult undertaking in practice, given that their optimal
relationship is user-dependent. In the present controlled experimental
scenario, the task is greatly eased since the size and performance of
the optimal solution is known in advance.  The aim of the experiments
is precisely to contrast the ability of the different FSAs to hit a
solution with respect to relevance, irrelevance,
redundancy and sample size.

\medskip\noindent \textbf{Relevance:} Different families of problems
are generated by varying the number of relevant features $N_R$. These
are features that will have an influence on the output and whose role
can not be assumed by any other subset.

\medskip\noindent \textbf{Irrelevance:} Irrelevant features are
defined as those not having any influence on the output. Their values
are generated at random for each example. For a problem with $N_{R}$
relevant features, different numbers of irrelevant features $N_I$ are
added to the corresponding data sets (thus providing with several
subproblems for each choice of $N_R$).

\medskip\noindent \textbf{Redundancy:} In this work, a redundancy
exists when a feature can take the role of another. Following a
parsimony principle, we are interested in the behaviour of the
algorithms in front of this simplest case.  If an algorithm fails to
identify redundancy in this situation (something that is actually
found in the experiments reported below), then this is interesting and
something we should be aware of.  This effect is obtained by choosing
a relevant feature randomly and replicating it in the data set. For a
problem with $N_{R}$ relevant features, different numbers of redundant
features $N_{R'}$ are added in a way analogous to the generation of
irrelevant features.

\medskip\noindent \textbf{Sample Size:} number of instances $|S|$ of a
data sample $S$. In these experiments, $|S| = \alpha k N_T c$, where
$\alpha$ is a constant, $k$ is a multiplying factor, $N_T$ is the
total number of features ($N_{R} + N_I + N_{R'}$) and $c$ is the
number of classes of the problem. This means that the sample size will
depend linearly on the total number of features.

\subsection{Evaluation of performance}
We derive in this section a \emph{scoring} measure to capture the
degree to which a solution obtained by a FSA matches (one of) the
correct solution(s). This criterion behaves as a {\em similarity} $s:
{\cal P}(X) \times {\cal P}(X) \rightarrow [0, 1]$, between subsets of
$X$ in the data analysis sense \citep{Chanpin:1973}, where $s(X_1,X_2)
> s(X_1,X_3)$ indicates that $X_2$ is more similar to $X_1$ than
$X_3$, and satisfying $s(X_1,X_2) = 1 \Longleftrightarrow X_1=X_2$ and
$s(X_1,X_2) = s(X_2,X_1)$. Let us denote by $X$ the total set of
features, partitioned in $X= X_R \cup X_I \cup X_{R'}$, being $X_R,
X_I, X_{R'}$ the subsets of relevant, irrelevant and redundant
features of $X$, respectively and call $X^* \subseteq X$ any of the
correct solutions (all and only relevant variables, no redundancy).
Let us denote by ${\cal A}$ the feature subset selected by a FSA. The
idea is to check how much ${\cal A}$ and $X^*$ have in common.

Let us define ${\cal A}_R = X_R \cap {\cal A}$, ${\cal A}_I = X_I \cap
{\cal A}$ and ${\cal A}_{R'} = X_{R'} \cap {\cal A}$. In general, we
have ${\cal A}_T = X_T \cap {\cal A}$ (hereafter $T$ stands for a
subindex in $\{R, I, R'\}$). Since necessarily ${\cal A} \subseteq X$,
we have that ${\cal A}={\cal A}_R \cup {\cal A}_I \cup {\cal A}_{R'}$
is a partition of ${\cal A}$. The \emph{score} $S_X({\cal A}): {\cal
  P}(X) \rightarrow [0, 1]$ is defined in terms of the similarity in
that for all ${\cal A} \subseteq X, S_X({\cal A}) = s({\cal A},X^*)$.
Thus, $S_X({\cal A}) > S_X({\cal A}')$ indicates that ${\cal A}$ is
more similar to $X^*$ than ${\cal A}'$. The idea is to make a flexible
measure, so that it can ponder each type of divergence (in relevance,
irrelevance and redundancy) to the correct solution. To this end, a
set of parameters is collected as $\alpha = \{\alpha_R, \alpha_I,
\alpha_{R'}\}$ with $\alpha_T \geq 0$ and $\sum \alpha_T = 1$.

\medskip\noindent
\textbf{Intuitive Description.}
The criterion $S_X({\cal A})$ penalizes three situations: (1) There
are relevant features lacking in ${\cal A}$ (the solution is
\textit{incomplete}), (2) There are more than enough relevant features
in ${\cal A}$ (the solution is \textit{redundant}) and (3) There are
some irrelevant features in ${\cal A}$ (the solution is
\textit{incorrect}).

An order of importance and a weight will be assigned (via the
$\alpha_T$ parameters), to each of these situations.  The precedent
point (3) is simple to model: if suffices to check whether $|{\cal
  A}_I| > 0$, being ${\cal A}$ the solution of the FSA. Relevance and
redundancy are strongly related given that a feature is redundant or
not depending on what other relevant features are present in ${\cal
  A}$.  Notice then that the correct solution $X^*$ is not unique, and
all of them should be equally valid. To this end, the features are
broken down in \emph{equivalence classes}, where elements of the same
class are redundant to each other (i.e., any correct solution must
comprise only one feature of each equivalence class). Being $A$ a
feature set, we define a binary relation between two features $x_i,
x_j \in A$ as: $x_i \thicksim x_j \Longleftrightarrow$ $x_i$ and $x_j$
represent the same information.  Clearly $\thicksim$ is an equivalence
relation. Let $A/{\thicksim}$ be the quotient set of $A$ under
$\thicksim$; any correct solution must be of the same size than $X_R$
and have one element in every subset of $(X_R \cup
X_{R'})/{\thicksim}$.

\medskip\noindent \textbf{Construction of the \emph{score}.}  The set
to be split in equivalence classes is formed by all the relevant
features (redundant or not) chosen by a FSA. Define ${\rho}_{\cal A} =
({\cal A}_R \cup {\cal A}_{R'})/{\thicksim}$ {\it (equivalence classes
  in which the relevant and redundant features chosen by a FSA are
  split)}, ${\rho}_X = (X_R \cup X_{R'})/{\thicksim}$ {\it (same with
  respect to the original set of features)} and ${\rho}_{{\cal A}
  \subseteq X} = \{ x \in {\rho}_X\ |\ \exists y \in {\rho}_{\cal A},
y \subseteq x \}$. For $Q$ quotient set, let:

$$ F(Q) = \sum_{x \in Q} (|x|-1) $$

The idea is to express the \emph{quotient} between the number of
redundant features chosen by the FSA and the number it could have
chosen, given the relevant features present in its solution.
In the precedent notation, this is written (provided the
denominator is not null):

$$ \frac{F({\rho}_{\cal A})}{F({\rho}_{{\cal A} \subseteq X})} $$

Let us finally build the \emph{score}, formed by three terms:
relevance, irrelevance and redundancy. Defining:

$$ I_{\cal A} = 1 - \frac{|{\cal A}_I|}{|X_I|}, \qquad R_{\cal A} = \frac{|{\rho}_{\cal A}|}{|X_R|},$$
$$
 R'_{\cal A} = \left\{
 \begin{array}{ll}
    0 & \textrm{if  } F({\rho}_{{\cal A} \subseteq X}) = 0 \\
    \left(1-\frac{F({\rho}_{\cal A})}{F({\rho}_{{\cal A} \subseteq X})}\right) &
    \textrm{otherwise.}
  \end{array}
 \right.
   $$

\medskip

\noindent for any ${\cal A} \subseteq X$ the score is defined as
$S_X({\cal A}) = s({\cal A},X^*) = \alpha_R R_{\cal A} + \alpha_{R'}
R'_{\cal A} + \alpha_I I_{\cal A}$. This score fulfills the two
conditions (proof is given
in the Appendix): 

\begin{enumerate}
	\item $S_X({\cal A}) = 0 \Longleftrightarrow {\cal A} = X_I$
  \item $S_X({\cal A}) = 1 \Longleftrightarrow {\cal A} = X^*$ 
\end{enumerate}

We can establish now the desired restrictions
on the behavior of the score. From the more to the less severe: there
are relevant features lacking, there are irrelevant features, and
there is redundancy in the solution. This is reflected in the
following conditions on the $\alpha_T$:

\begin{enumerate}
  \item Choosing an irrelevant feature is better than missing a relevant one:
  $\frac{\alpha_R}{|X_R|} > \frac{\alpha_I}{|X_I|}$
  \item Choosing a redundant feature is better than choosing an
irrelevant one:
  $\frac{\alpha_I}{|X_I|} > \frac{\alpha_{R'}}{|X_{R'}|}$
\end{enumerate}

We also define $\alpha_T = 0$ if $|X_T| = 0$. Observe that the
denominators are important for, say, expressing the fact that it is
not the same choosing an irrelevant feature when there were only two
that when there were three (in the latter case, there is an irrelevant
feature that could have been chosen when it was not).  In order to
translate the previous inequalities into workable conditions, a
parameter $\epsilon \in (0,1]$ is introduced to express the precise
relation between the $\alpha_T$. Let $\underline{\alpha}_T =
\frac{\alpha_T}{|X_T|}$. The following equations have to be satisfied,
together with ${\alpha}_R + {\alpha}_I + {\alpha}_{R'} = 1$:

$$ \beta_R \underline{\alpha}_R = \underline{\alpha}_I , \qquad
\beta_I \underline{\alpha}_I = \underline{\alpha}_{R'} $$ 

for suitable chosen values of $\beta_R$ and $\beta_I$. Reasonable
settings are obtained by taking $\beta_R = \epsilon / 2$ and $\beta_I
= 2\epsilon / 3$, though other settings are possible, depending on the
evaluator's needs. With these values, at equal $|X_R|, |X_I|, |X_{R'}|$,
$\alpha_R$ is at least twice more important than $\alpha_I$ (because
of the $\epsilon / 2$) and $\alpha_I$ is at least one and a half times
more important than $\alpha_{R'}$.  Specifically, the minimum values
are attained for $\epsilon=1$ (i.e., $\alpha_R$ counts twice
$\alpha_I$). For $\epsilon < 1$ the differences widen proportionally
to the point that, for $\epsilon \approx 0$, only $\alpha_R R$ will
practically count on the overall score.

\section{EXPERIMENTAL EVALUATION}
\label{section:exp-eval}

In the following sections we detail the experimental methodology and quantify the various
parameters of the experiments. The basic idea consists on generating sample data sets using
synthetic functions $f$ with known relevant features. These data sets (of different sizes) are
corrupted with irrelevant and/or redundant features and handed over to the different FSAs to
obtained a hypothesis  $H$. The divergence between the defined function $f$ and the obtained
hypothesis $H$ will be evaluated by the \emph{score} criterion (with $\epsilon=1$). This
experimental design is illustrated in Fig. \ref{fig:generalview}.

\subsection{Description of the FSAs used}
\label{section:algoritmos2}

Up to ten FSAs were used in the experiments. These are \textsc{E-Sfg},
\textsc{Qbb}, \textsc{Lvf}, \textsc{Lvi}, \textsc{C-Sbg},
\textsc{Relief}, \textsc{Sfbg}, \textsc{Sffg}, \textsc{W-Sbg}, and
\textsc{W-Sfg}. The algorithms \textsc{E-Sfg}, \textsc{W-Sfg} are
versions of \textsc{Sfg} using entropy and the accuracy of a C4.5
inducer, respectively. The algorithms \textsc{C-Sbg}, \textsc{W-Sbg}
are versions of \textsc{Sbg} using consistency and the accuracy of a
C4.5 inducer, respectively. Since \textsc{Relief} and \textsc{E-Sfg}
yield an ordered list of features $x_i$ according to their weight
$w_i$, an automatic filtering criterion is necessary to transform
every solution into a subset of features. The procedure used here to
determine a suitable cut point is simple: first the weights are sorted
in decreasing order (with $w_n$ the greatest weight, corresponding to
the most relevant feature). Then those weights further than two
variances from the mean are discarded (that is to say, with very high
or very low weights). The idea is to look for the feature $x_j$ such
that $\frac{w_n - w_j}{w_n - w_1} \frac{j}{n} \quad \mbox{is
  maximum.}$ Intuitively, this corresponds to obtaining the maximum
weight with the lowest number of features.  The cut point is then set
between $x_j$ and $x_{j-1}$.

\setcounter{figure}{0}
\renewcommand{\figurename}{Figure}

\begin{figure}[bt!]
\centering
\includegraphics[width=7cm]{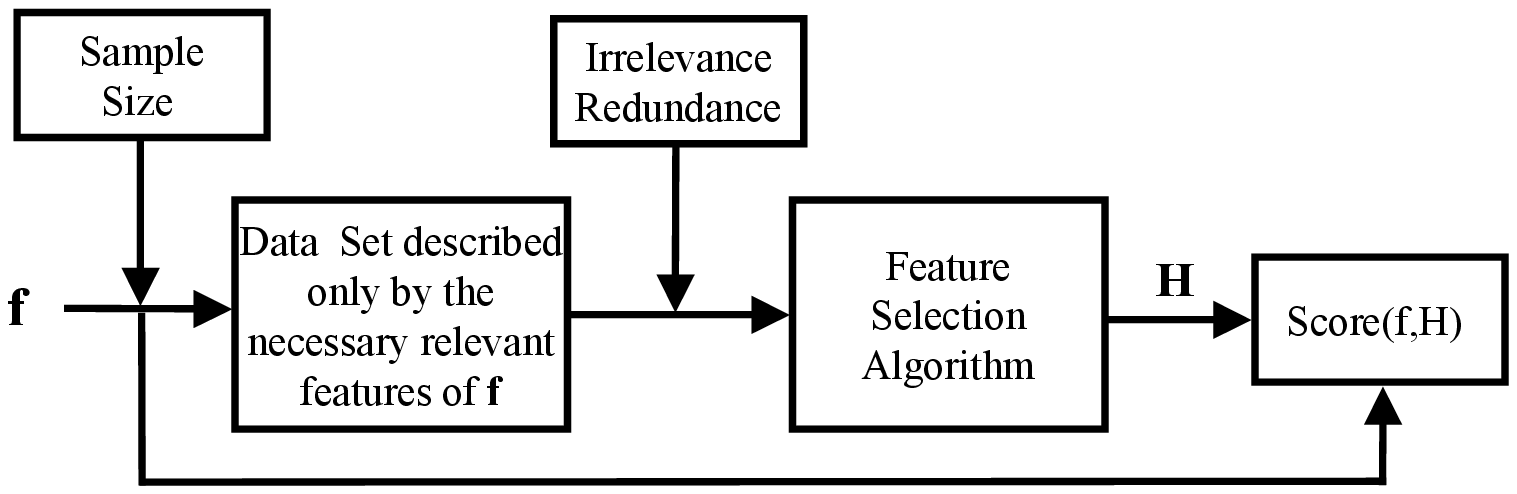}
\caption{Flow-Chart of Experimental Design.}
\label{fig:generalview}
\end{figure}

\subsection{Implementations of data families} \label{section:datos}

A total of twelve families of data sets were generated studying three
different problems and four instances of each, by varying the number
of relevant features $N_R$. Let $x_1,\ldots,x_n$ be the relevant
features of a problem $f$.

\medskip\noindent {\bf Parity:}
This is the classic problem where the output is $f(x_1, \cdots, x_n)=1$ if the number of $x_i=1$
is odd and $f(x_1, \cdots, x_n)=0$ otherwise.

\medskip\noindent
{\bf Disjunction:} Here we have $f(x_1, \cdots, x_n) = 1$ if $(x_1 \wedge \cdots \wedge x_{n'})
\vee (x_{n'+1} \wedge \cdots \wedge x_{n})$, with $n'=n\, div\, 2$  ($n$ even) and $n'=(n\, div\,
2) + 1$ ($n$ odd).

\medskip\noindent
{\bf GMonks:} This problem is a generalization of the classic \emph{monks} problems
\citep{Thrun:1991}. In its original version, three independent problems were applied on sets  of
$n=6$ features that  take values of a discrete, finite and unordered set (nominal features). Here
we have grouped the three problems in a single one computed on  \emph{each} chunk of 6 features.
Let $n$ be multiple of 6, $k = n \, div\, 6$ and $b = 6(k'-1)+1$, for $1 \leq k' \leq k$. Let us
denote for ``1'' the first value of a feature, for ``2'' the second, etc. The problems are the
following:

\begin{enumerate}
  \item $P1: (x_b = x_{b+1}) \vee x_{b+4} = 1$
  \item $P2:$ two or more $x_i = 1$ in $x_b \cdots x_{b+5}$
  \item $P3: (x_{b+4} = 3 \wedge x_{b+3} = 1) \vee (x_{b+4} \neq 3 \wedge x_{b+1} \neq 2)$
\end{enumerate}

For each chunk, the boolean condition $P2 \wedge \neg (P1 \wedge P3)$ is checked. If it is
satisfied for $n_c\, div\, 2$ or more chunks (being $n_c$ the number of chunks) the function
\emph{Gmonks} is 1; otherwise, it is 0.

\subsection{Experimental setup}
The experiments are divided in three main groups. The first group
explores the relationship between \emph{irrelevance vs. relevance}.
The second one explores the relationship between \emph{redundancy vs.
  relevance}. The last group is the study of the effect of different
\emph{sample sizes}. Each group uses three families of problems
(\emph{Parity}, \emph{Disjunction} and \emph{GMonks}) with four
different instances for each one, varying the number of relevant
features $N_R$, as indicated:

\medskip\noindent\textbf{Relevance:} The different numbers $N_R$ vary
for each problem, as follows: \{4, 8, 16, 32\} (for \emph{Parity}),
\{5, 10, 15, 20\} (for \emph{Disjunction}) and \{6, 12, 18, 24\} (for
\emph{GMonks}).

\noindent\textbf{Irrelevance:} In these experiments, $N_I$
runs from zero to twice the value of $N_R$. Specifically, $N_I \in \{
(k \cdot N_R)/p, k=0,1,\ldots,10\}$ (that is, eleven different
experiments of irrelevance for each $N_R$). The value of $p$ is chosen
so that all the involved quantities are integer: $p=4$ for
\emph{Parity}, $p=5$ for \emph{Disjunction} and $p=6$ for
\emph{GMonks}.

\noindent\textbf{Redundancy:} Analogously to the
generation of irrelevant features, we have $N_{R'}$ running from zero
to twice the value of $N_{R}$ (eleven experiments of irrelevance for
each $N_R$).

\noindent\textbf{Sample Size:} Given the formula $|S| = \alpha k N_T c$ (see
\S\ref{section:experimental-designer}), different problems were
generated considering $k \in \{$0.25, 0.5, 0.75, 1.0, 1.25, 1.75,
2.0$\}$, $N_T = N_R + N_I + N_{R'}$, $c=2$ and $\alpha = 20$.  The
values of $N_I$ and $N_{R'}$ were fixed as $N_I = N_{R'} = N_R\, div\,
2$.

\subsection{Discussion of the results}

\begin{figure*}[htb!]
 \begin{center}

\begin{tabular}{c@{\hspace{1pc}}c}

 \includegraphics[width=7.5cm,height=6.0cm]{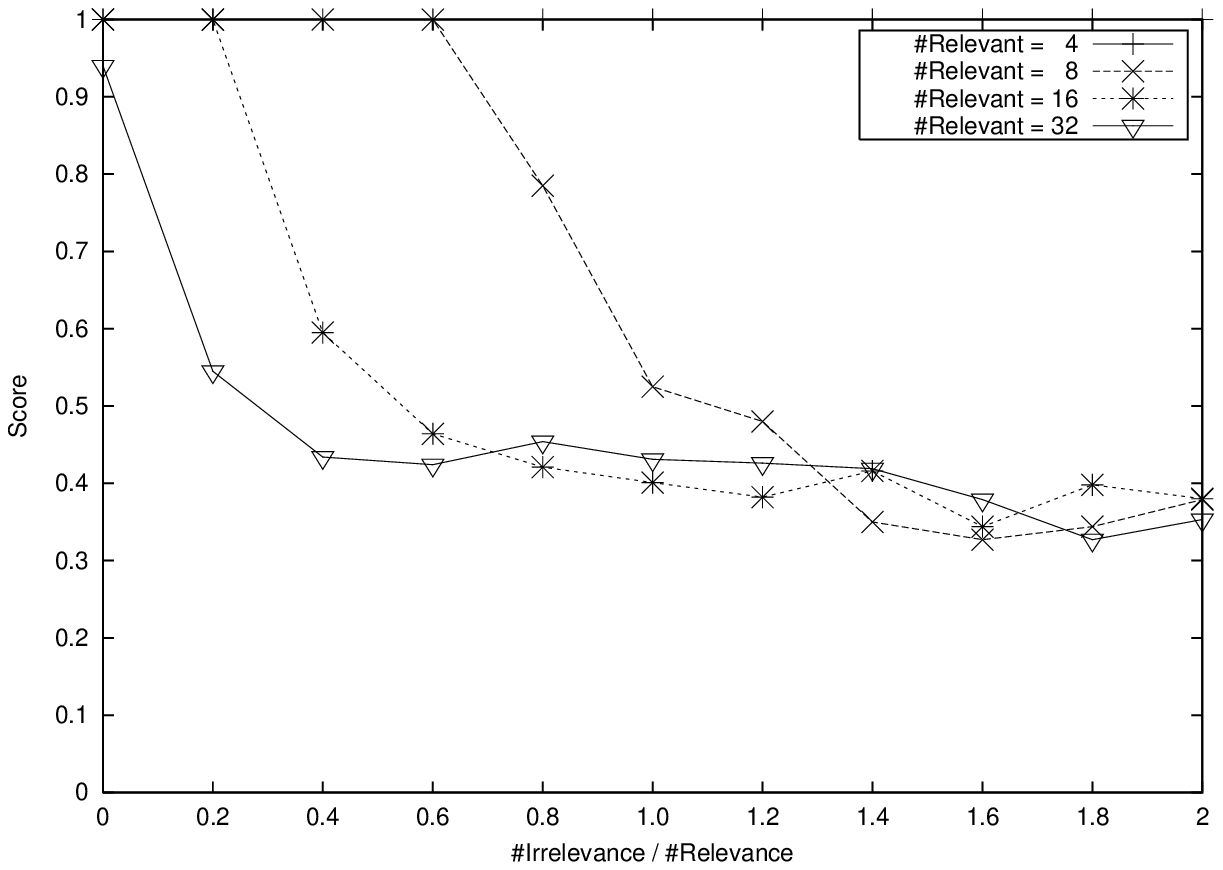} &
 \includegraphics[width=7.5cm,height=6.0cm]{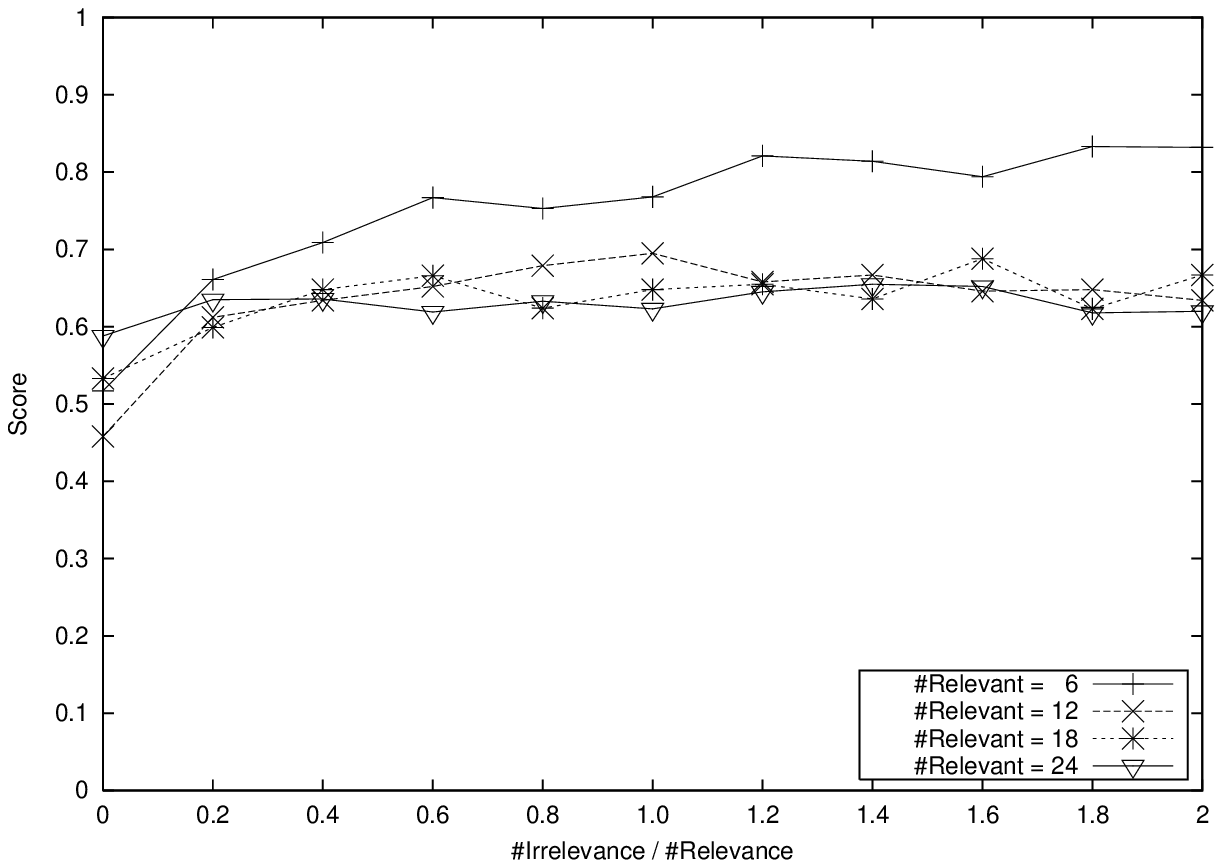} \\
  a. Irrelevance vs. Relevance\\for \textit{Parity} with \textsc{C-Sbg} & b. Irrelevance vs. Relevance\\&for \textit{GMonks} with \textsc{Relief} \\

 \includegraphics[width=7.5cm,height=6.0cm]{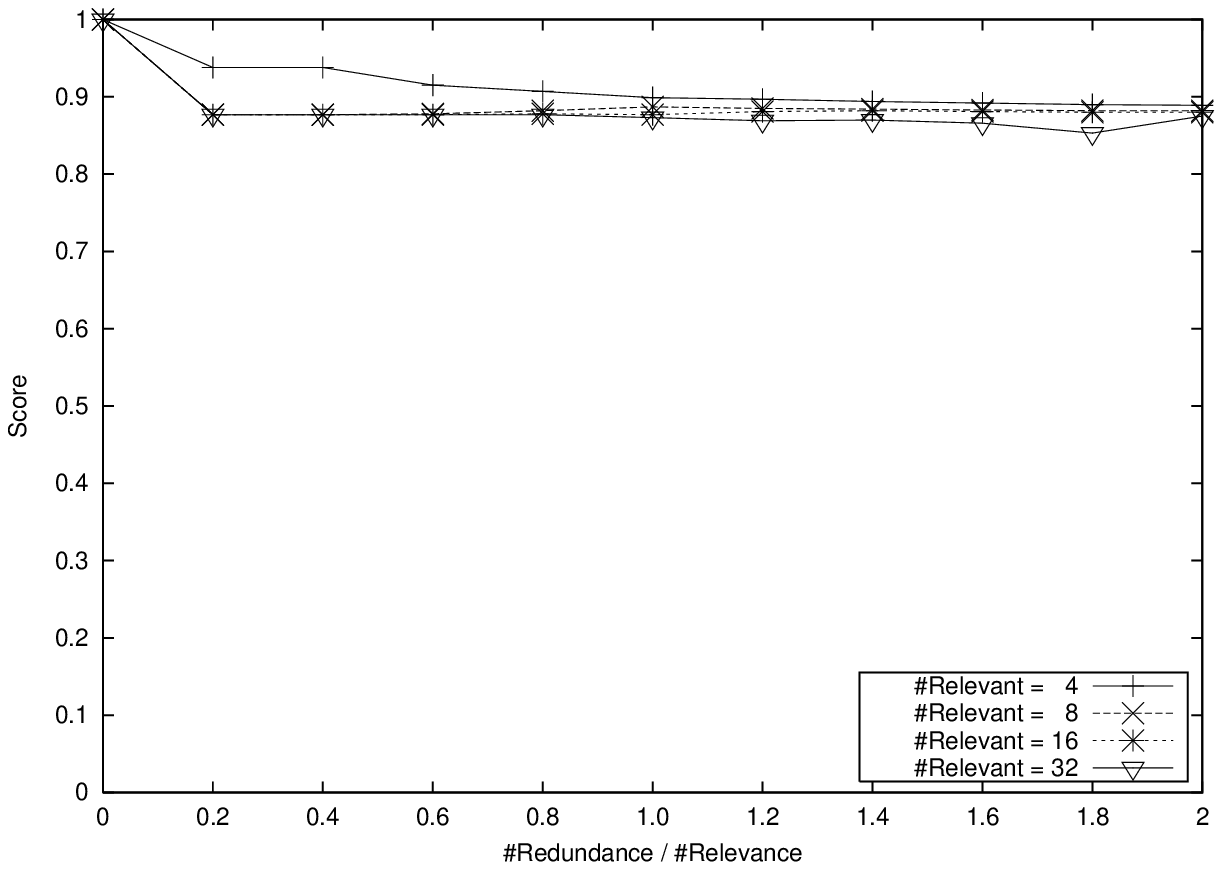} &
 \includegraphics[width=7.5cm,height=6.0cm]{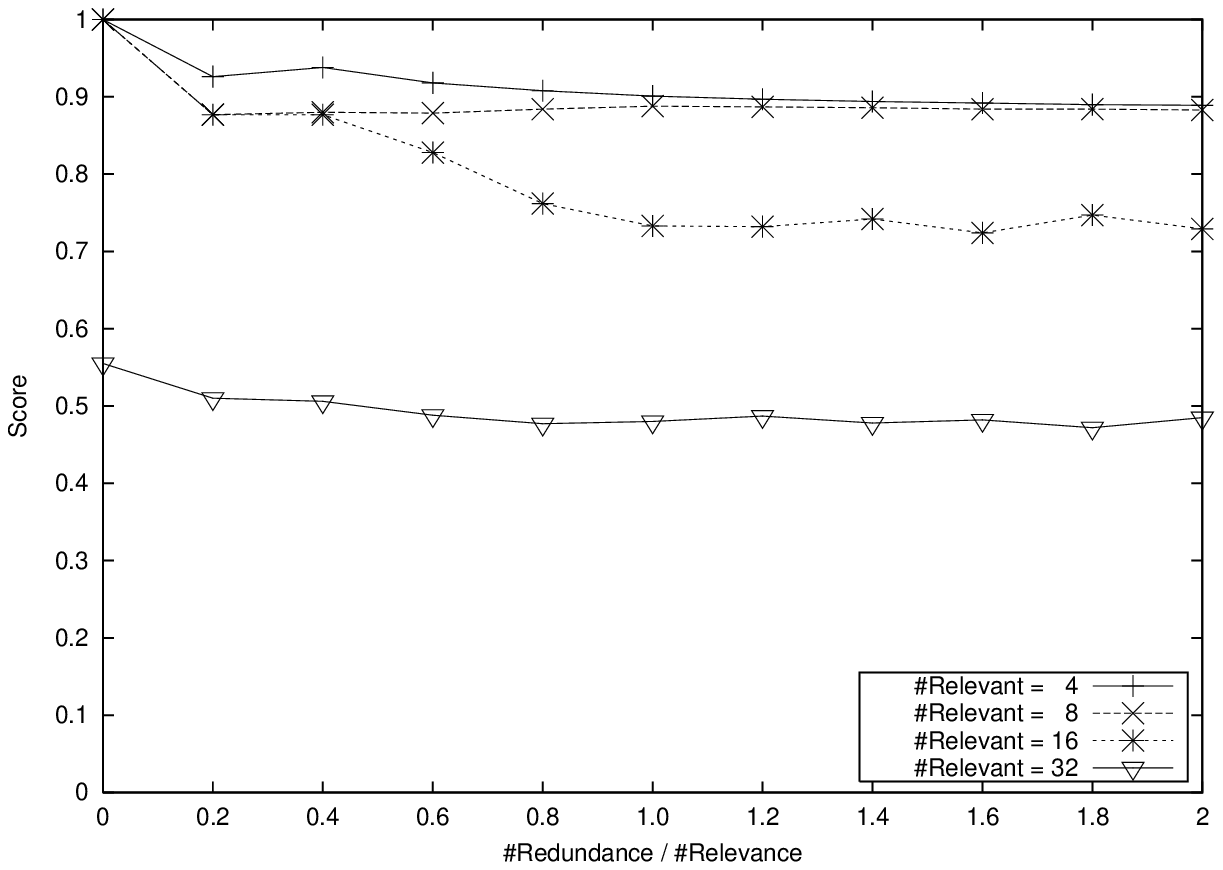} \\
 c. Redundance vs. Relevance\\for \textit{Parity} with \textsc{Lvf} & d. Redundance vs. Relevance\\&for \textit{Disjunction} with \textsc{Qbb} \\

 \includegraphics[width=7.5cm,height=6.0cm]{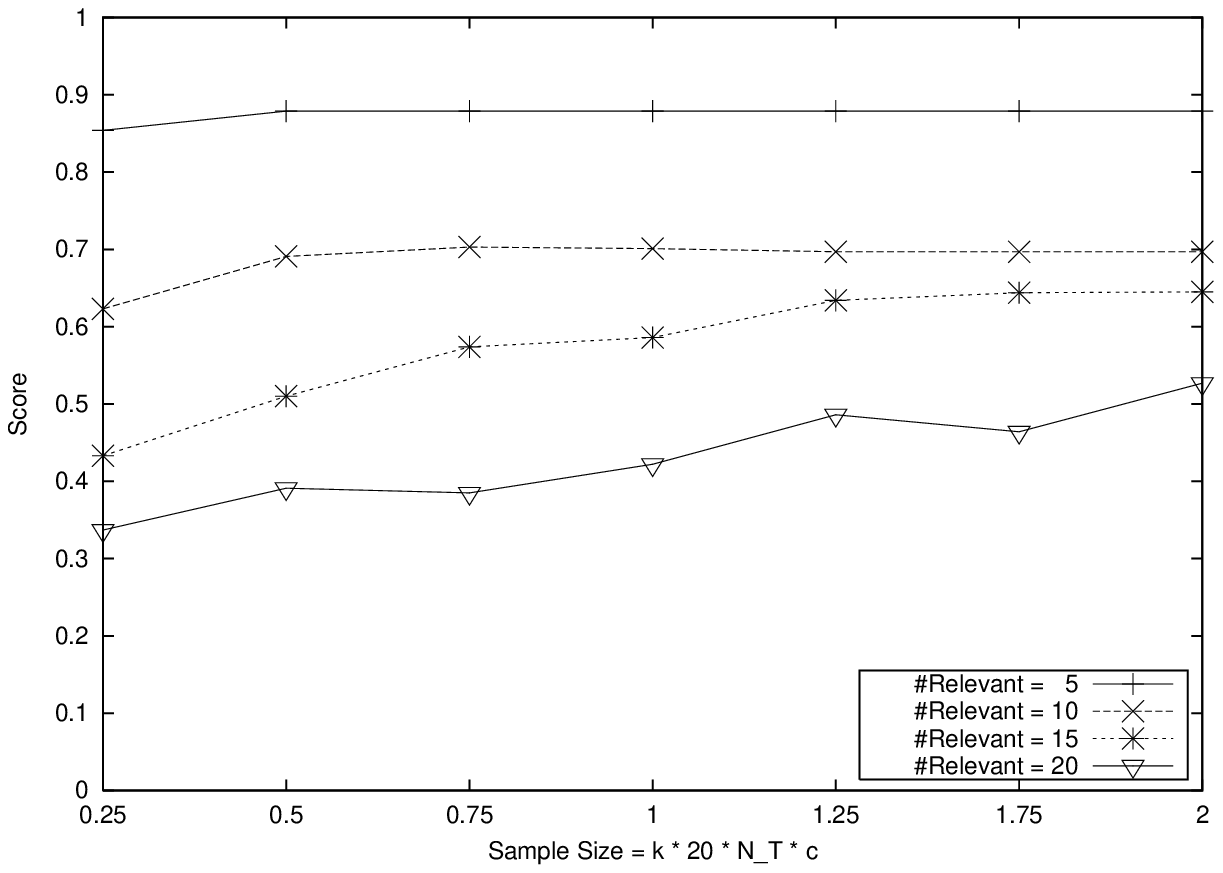} &
 \includegraphics[width=7.5cm,height=6.0cm]{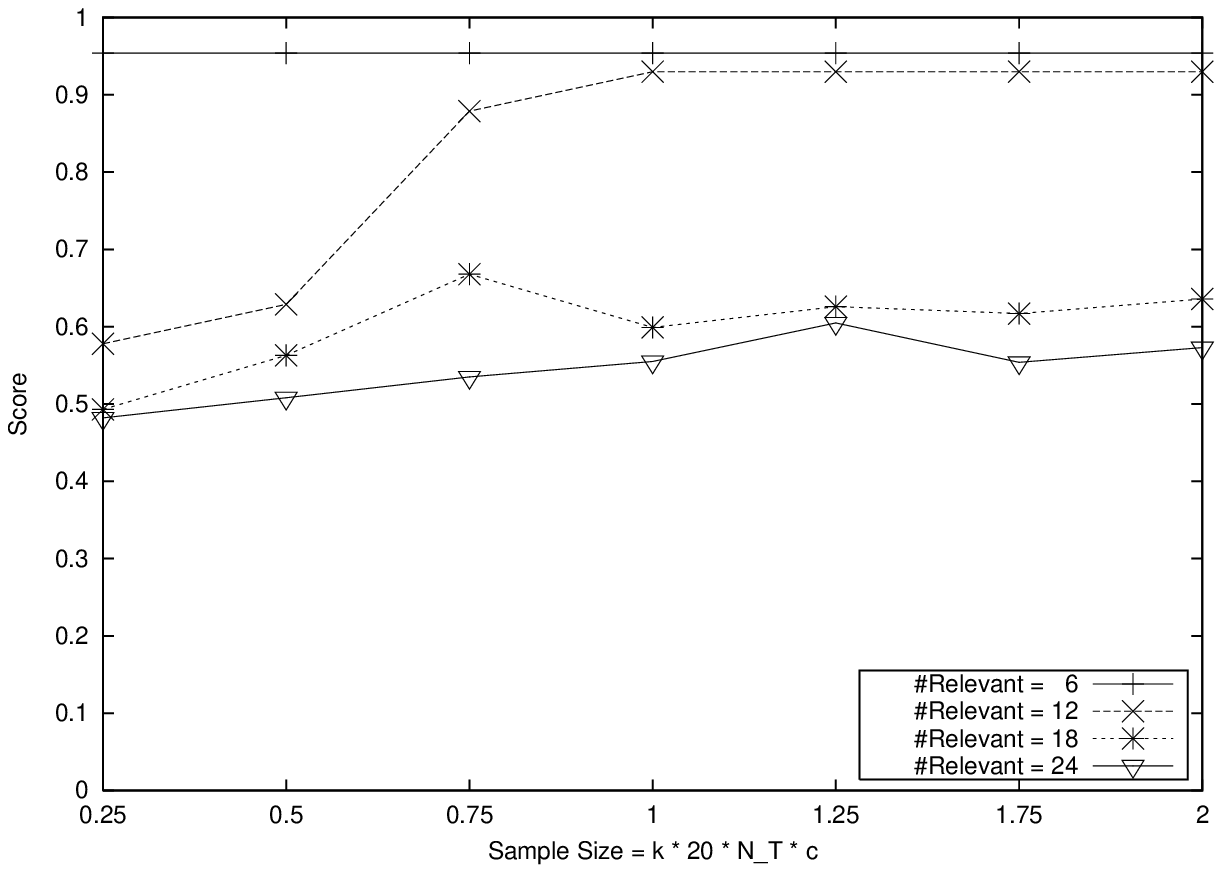} \\
 e. Sample Size\\for \textit{Disjunction} with \textsc{Lvi} & f. Sample Size\\&for \textit{Parity} with \textsc{W-Sbg} \\
\end{tabular}
\end{center}

\caption{Selected results of the experiments:
  (a),(b) is {\em irrelevance} vs. {\em relevance}, (c),(d) are examples of
  {\em redundancy} vs. {\em relevance} and (e), (f) of {\em sample size}
  experiments. The horizontal axis is the {\em ratio} between these
  quantities. The vertical axis is the average result given by the
  \textit{score} in 10 independent runs with different random data samples.}
\label{fig:experimentos}
\end{figure*}

\begin{figure*}[htb!]
 \begin{center}

\begin{tabular}{c@{\hspace{1pc}}c}

 \includegraphics[width=7.5cm,height=7.0cm]{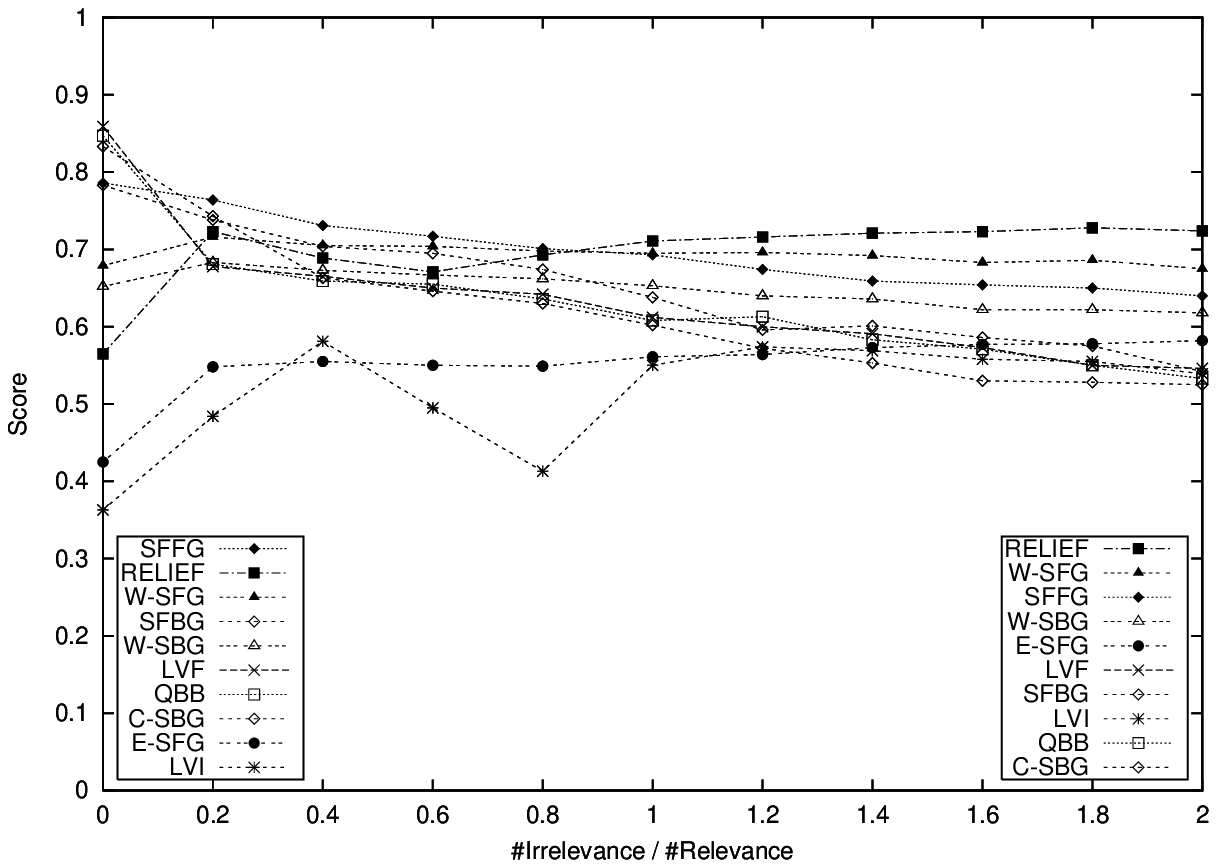} &
 \includegraphics[width=7.5cm,height=7.0cm]{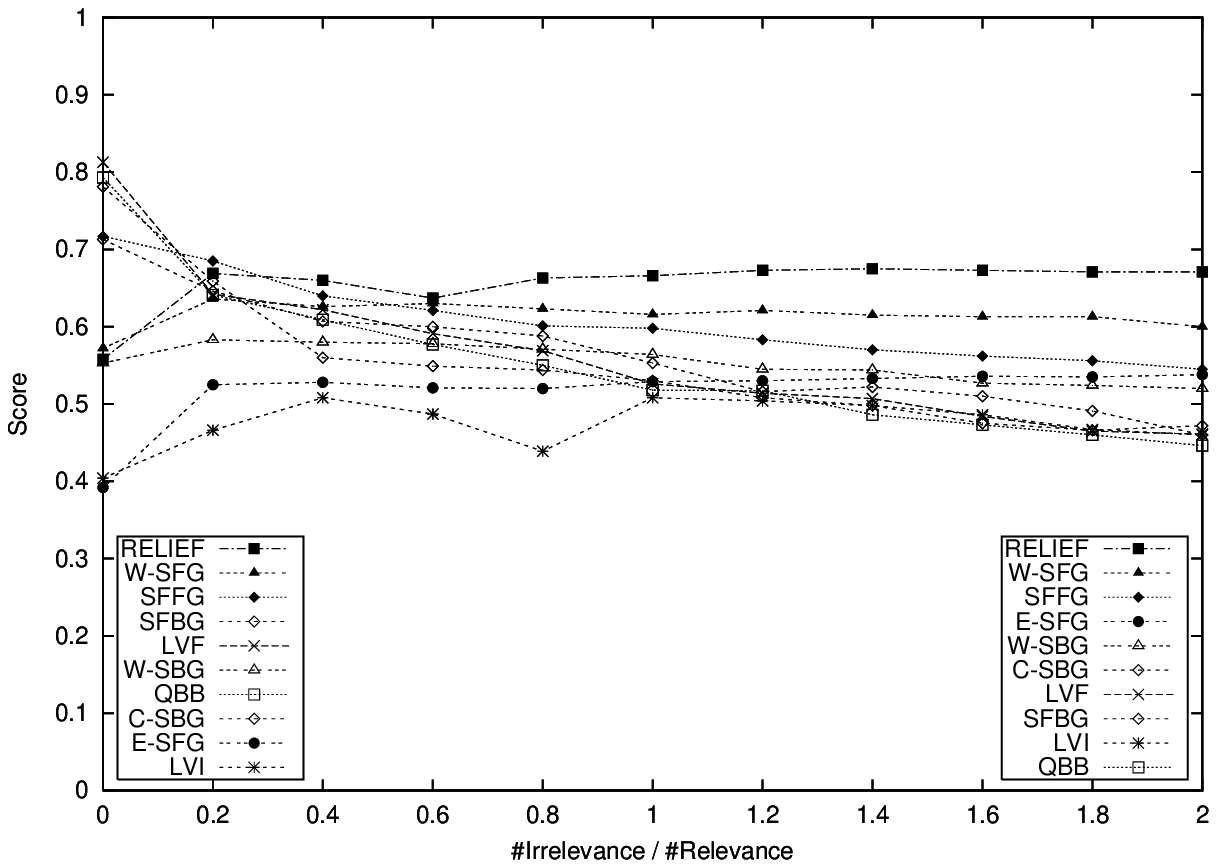} \\
  a. Irrelevance  & b. Irrelevance (weighed) \\

 \includegraphics[width=7.5cm,height=7.0cm]{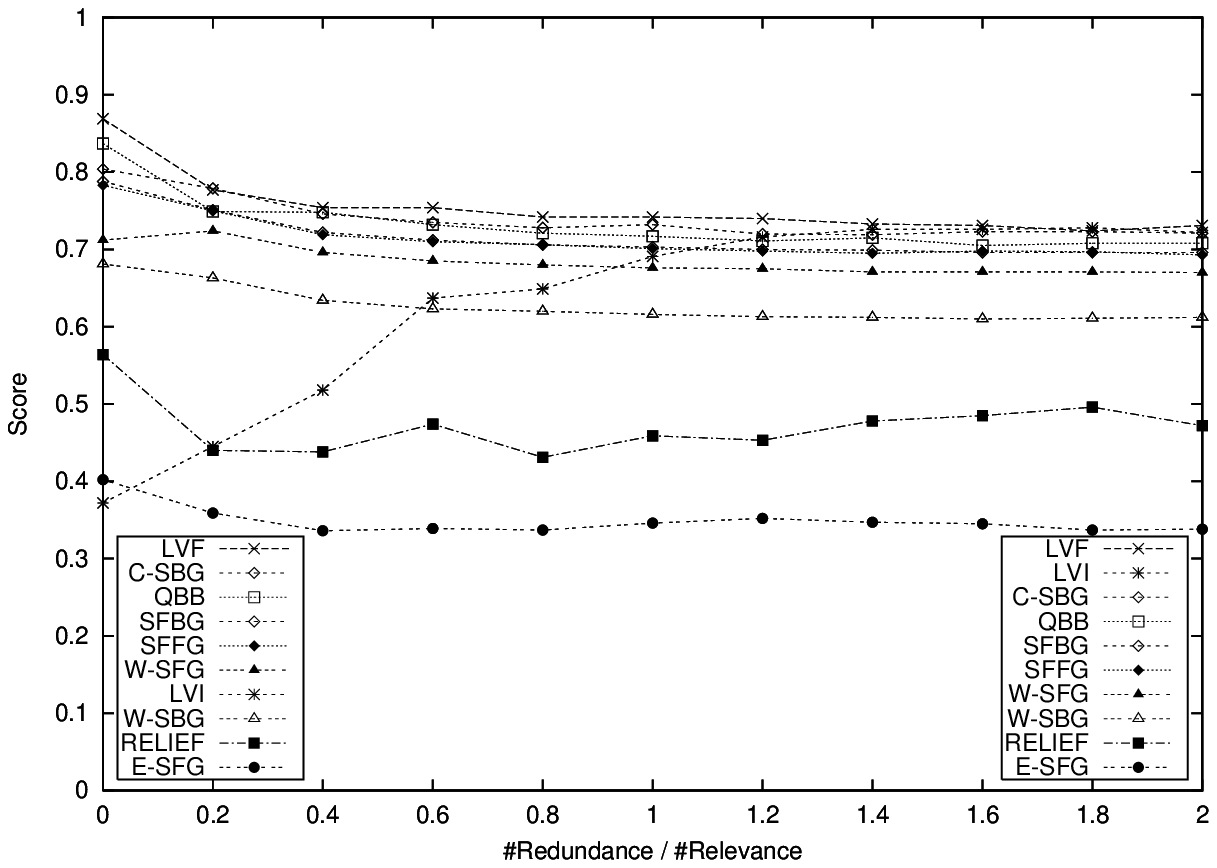} &
 \includegraphics[width=7.5cm,height=7.0cm]{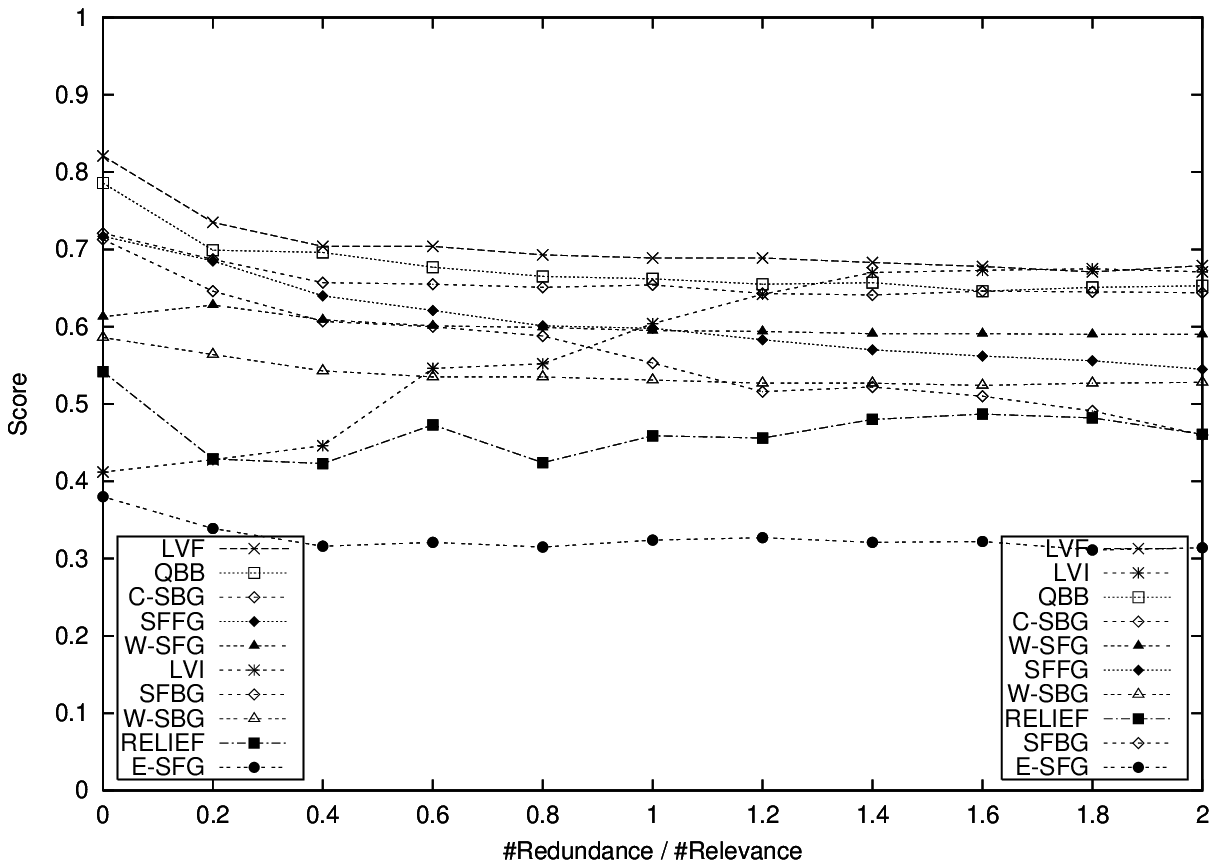} \\
  c. Redundance  & d. Redundance (weighed) \\

 \includegraphics[width=7.5cm,height=7.0cm]{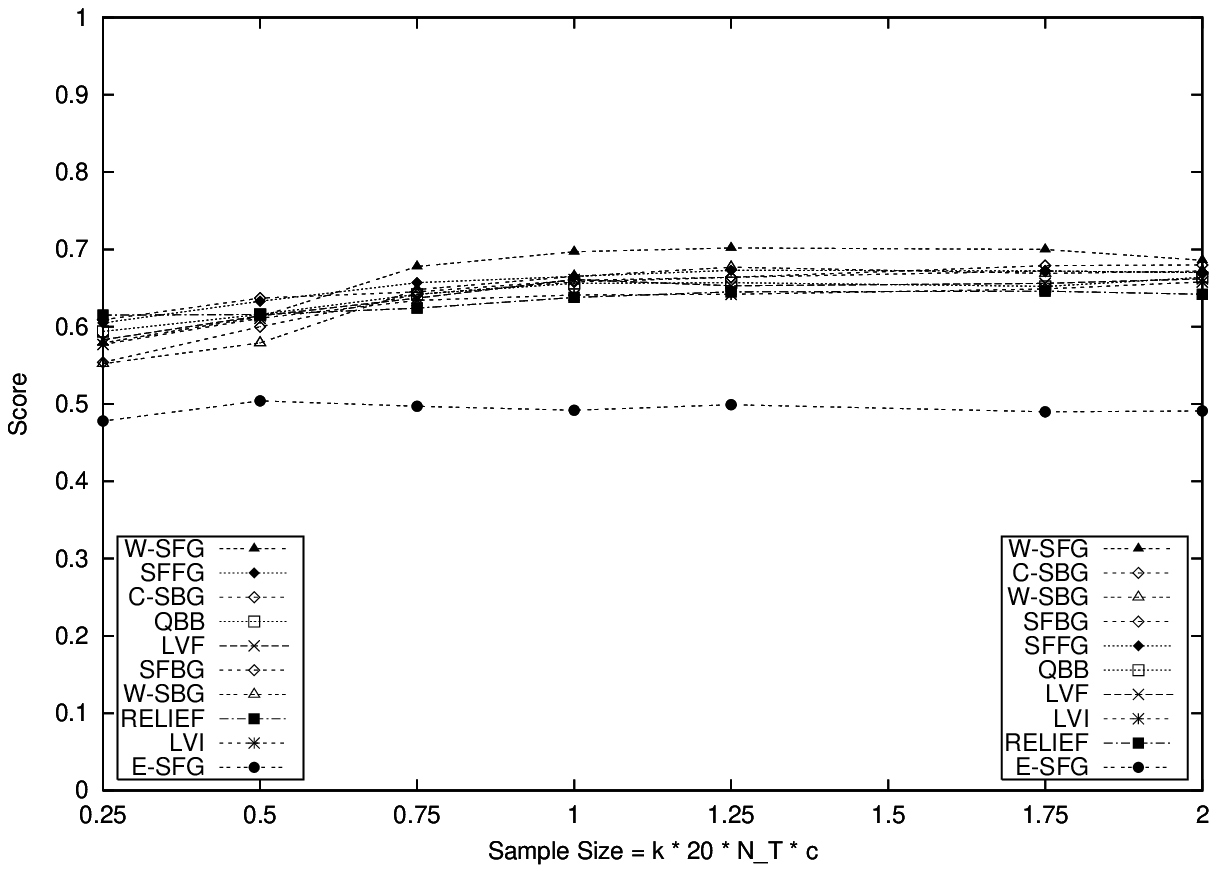} &
 \includegraphics[width=7.5cm,height=7.0cm]{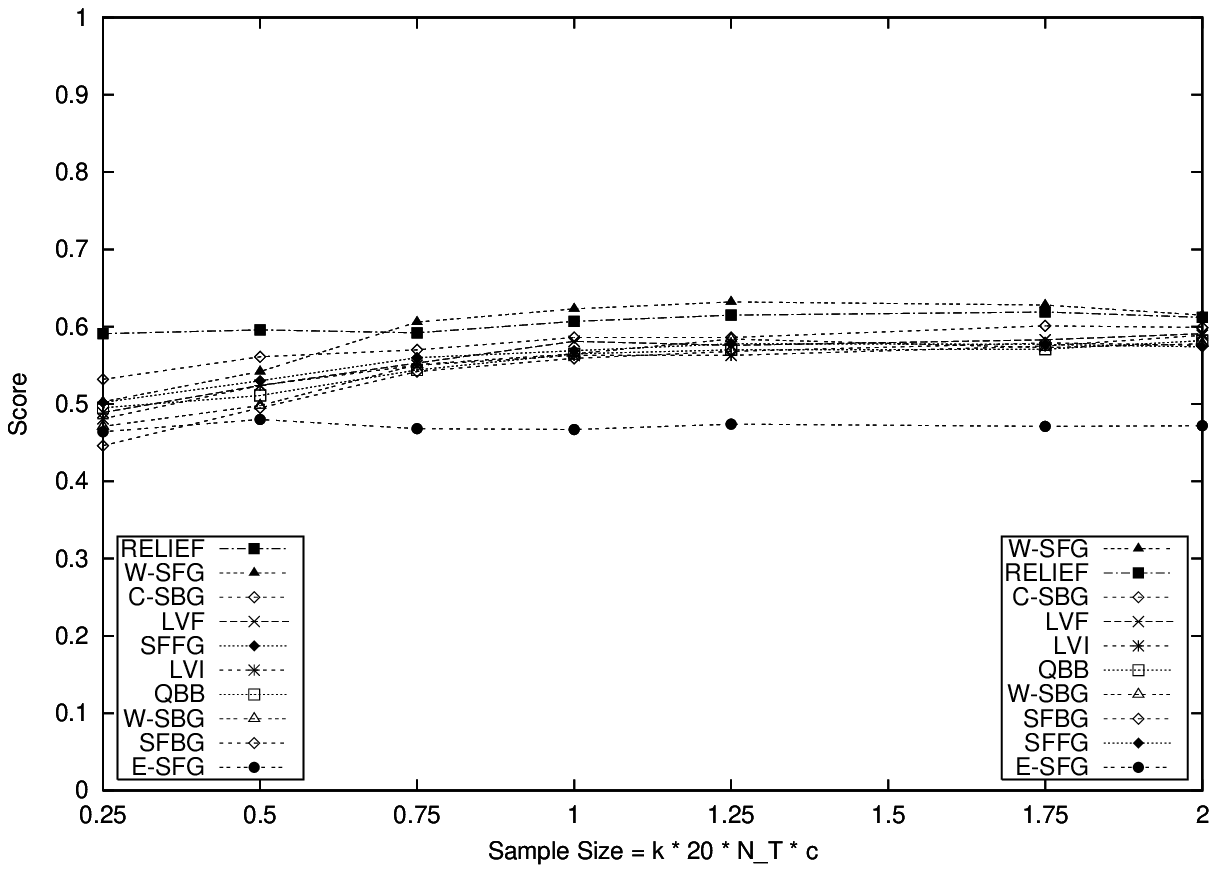} \\
  e. Sample Size  & f. Sample Size (weighed) \\

\end{tabular}
\end{center}

\caption{Results ordered by total average performance on the data
sets (left inset) and by end performance (right inset). Figs. (b),
(d) and (f) are weighed versions of (a), (c) and (e),
respectively.}\label{fig:experimentos2}
\end{figure*}

Due to space reasons, only a representative sample of the results is
presented, in graphical form, in Figs. \ref{fig:experimentos} and
\ref{fig:experimentos2}. In all the plots, each point represents the
average of 10 independent runs with different random data samples. The
Figs.  \ref{fig:experimentos}(a) and (b) are examples of {\em
  irrelevance vs.  relevance} for four instances of the problems, (c),
(d) are examples of {\em redundancy vs. relevance} and (e), (f) of
{\em sample size} experiments. In all cases, the horizontal axis
represents the {\em ratios} between these particulars as explained
above. The vertical axis represents the average results given by the
score criterion.

\begin{itemize}
  \item In Fig. \ref{fig:experimentos}(a) the \textsc{C-Sbg} algorithm
shows at first a good performance but clearly falls dramatically
(below the 0.5 level from $N_I=N_R$ on) as the {\em irrelevance ratio}
increases. Note that for $N_R=4$ performance is perfect
(the plot is on top of the graphic). In contrast, in
Fig.~\ref{fig:experimentos}(b) the \textsc{Relief} algorithm presents
very similar and fairly good results for the four instances of the
problem, being almost insensitive to the total number of features.

\item In Fig.~\ref{fig:experimentos}(c) the \textsc{Lvf} algorithm presents a
very good and stable performance for the different problem instances
of \emph{Parity}. In contrast, in \ref{fig:experimentos}(d)
\textsc{Qbb} tends to a poor general performance in the
\emph{Disjunction} problem when the total number of features increases.

\item The plots in Figs.~\ref{fig:experimentos}(e) and (f) show
  additional interesting results because we can appreciate the curse
  of dimensionality \citep{Jain:1997}. In these figures, \textsc{Lvi}
  and \textsc{W-Sfg} perform increasingly poorly (see the figure from
  top to bottom) with higher numbers of features, provided the number
  of examples is increased in a linear way. However, in general, as
  long as more examples are added, performance is better (left to
  right).
\end{itemize}

A summary of the \textit{complete} results is displayed in Fig.
\ref{fig:experimentos2} for the ten algorithms, allowing for a
comparison across all the sample datasets with respect to each studied
particular. Specifically, Figs.~\ref{fig:experimentos2}(a), (c) and
(d) show the average score of each algorithm for irrelevance,
redundancy and sample size, respectively. Moreover, Figs.
\ref{fig:experimentos2}(b), (d) and (f) show the same average weighed
by $N_R$, in such a way that more weight is assigned to more difficult
problems (higher $N_R$).  In each graphic there are two keys: the key
to the left shows the algorithms ordered by \emph{total} average
performance, from top to bottom. The key to the right shows the
algorithms ordered by average performance on the \emph{last} abscissa
value, also from top to bottom. In other words, the left list is
topped by the algorithm that wins on average, while the right list is
topped by the algorithm that ends on the lead. This is also useful to
help reading the graphics.

\begin{itemize}
  \item Fig.~\ref{fig:experimentos2}(a) shows that \textsc{Relief}
  ends up on the lead of the irrelevance vs. relevance problems,
  while \textsc{Sffg} shows the best average performance. The
  algorithm \textsc{W-Sfg} is also well positioned.

  \item Fig.~\ref{fig:experimentos2}(c) shows the algorithms
  \textsc{Lvf} and \textsc{Lvi}, together with \textsc{C-Sbg}, as
  the overall best. In fact, there is a bunch of algorithms that
  also includes the two \emph{floating} and \textsc{Qbb} showing a close
  performance. Note how \textsc{Relief} and the \emph{wrappers} are very
  poor performers.

\item Fig.~\ref{fig:experimentos2}(e) shows how the wrapper algorithms
  extract the most of the data when there is a shortage of it.
  Surprisingly, the backward wrapper is just fairly positioned on
  average. The \textsc{Sffg} algorithm is again quite good on average,
  together with \textsc{C-Sbg}. However, all of the algorithms are
  quite close and show the same kind of dependency to the amount of
  available data. Note the general poor performance of \textsc{E-Sfg},
  most likely due to the fact that it is the only algorithm that
  computes its evaluation measure (entropy in this case) independently
  for each feature.

\end{itemize}

The weighed versions of the plots (Fig.~\ref{fig:experimentos2}
(b),(d) and (f)) do not seem to alter the picture very much. A closer
look reveals that the differences between the algorithms have widened.
Very interesting is the change for \textsc{Relief}, that takes the
lead both on irrelevance and sample size, but not on redundancy.

\subsection{General considerations}
The results point to \textsc{Sffg} as the best algorithm on average in
complete ignorance of the particulars of the data set, or whenever one
is willing to use a single algorithm. However, in view of the reported
results, a better strategy would be to run various algorithms in a
\textit{coupled} way (i.e., in different execution orders and piping
the respective solutions) and observe the results. Specifically, we
suggest to use \textsc{Relief} when one is interested in detecting
{\em irrelevance}, \textsc{Lvf} for detecting {\em redundancy} and
\textsc{W-Sfg} in presence of small sample size situations.  In light
of this, we conjecture that \textsc{Sffg} used in a wrapper fashion
could be a better one-fits-all option for small to moderate size
problems.

\noindent
We would like to bring to attention the following points:

\begin{enumerate}
\item The wild differences in performance for different algorithms and
  data particulars: fixing an algorithm $A$ and a problem $P$,
  performance of $A$ is dramatically different for the various
  particulars considered (but in a consistent way in all instances of
  $P$). However, these results are coherent and scale quite well for
  increasing numbers of relevant features.
\item The \textit{score} criterion seems to reliably capture what intuition
tells about the quality of a solution at this simple level.
\end{enumerate}

We would also like to emphasize the fact that the differences in the
outcome yielded by the algorithms are not entirely due to their
different approach to the problem. Rather, they are also attributable
to the lack of a precise optimization goal, for example in the form
described in Definition \ref{def-fs}. Another good deal is the finite
(and possibly very limited) sample size which, on the one hand,
hinders the obtention of an accurate evaluation of relevance. On the
other, the dependence on a specific sample reminds us that every
evaluation of relevance in a feature subset should be regarded as the
outcome of a {\em random variable}, different samples yielding
different outcomes. In this vein, the use of {\em resampling}
techniques like Random Forests \citep{RF} is strongly recommended.

A final interesting point is the relation between the evaluation given
by a specific inducer and the score. We were interested in
ascertaining whether higher inducer evaluations imply higher
scores. We next provide evidence that this need not be the
case by means of a counterexample. 

{\em Conjecture}: given a FSA and the solution it yields in a data set, we
know this solution is suboptimal in the sense that better solutions
may exist but are not found. However, we would expect the solution to
be better (i.e. have a higher score) the better its performance is.

{\em Experiment}: we run \textsc{W-Sfg} in 10 independent runs with
different random data samples of size 600 using Na\"{\i}ve Bayes as inducer
in an instance of the {\em GMonks} problem, described by
$N_R = 24, N_{R'} = 12$ and $N_I = 24$ for $N_T = 60$. Table \ref{Ads}
shows the results: for each run, the final inducer performance is
given, as well as the score of the solutions. Runs 5 and
8 correspond to very different solutions (number 5 being much better
than number 8) that have almost the same inducer evaluation. Run 5
also has a lower evaluation than run 9, but a greater score.

\begin{table}[!hbt]
\centering
\begin{tabular}{|r|c|c|c|c|}\hline
\# & Na\"{\i}ve Bayes  & score $s$ & $|{\cal A}_R \cup {\cal A}_{R'}|$ & $|{\cal A}_I|$\\\hline
1& 0.876  & 0.603  & 22 & 15 \\\hline
2& 0.869  & 0.601  & 24 & 14 \\\hline
3& 0.884  & 0.588  & 19 & 10 \\\hline
4& 0.858  & 0.609  & 22 & 13 \\\hline
5& 0.876  & 0.730  & 30 & 19 \\\hline
6& 0.875  & 0.475  &  5 &  0 \\\hline
7& 0.872  & 0.456  &  8 &  6 \\\hline
8& 0.880  & 0.412  &  5 &  2 \\\hline
9& 0.881  & 0.630  & 14 &  4 \\\hline
10& 0.873 & 0.630  & 28 & 20 \\\hline
\end{tabular}
\caption{Results of the experiment on variability.}
\label{Ads}
\end{table}

This same experiment can be used to show the variability in the
results as a function of the data sample. It can be seen that the
numbers of relevant and redundant as well as irrelevant features
depend very much on the sample. A look at the precise features chosen
reveals that they are very different solutions (a fact that is also
indicated by the score) that nonetheless give a similar evaluation by
the inducer. Given the incremental nature of \textsc{W-Sfg}, it can
be deduced that classifier improvements where obtained by adding
completely irrelevant features.


\section{CONCLUSIONS}

The task of a \textit{feature selection algorithm} (FSA) is to provide
with a computational solution to the feature selection problem
motivated by a certain definition of \emph{relevance} or, at least, by
a performance evaluation measure.  This algorithm should also be
increasingly reliable with sample size and pursue the solution of a
clearly stated optimization goal. The many algorithms proposed in the
literature are based on quite different principles and loosely follow
these recommendations, if at all. In this research, several
fundamental algorithms have been studied to assess their performance
in a controlled experimental scenario. A measure to evaluate FSAs has
been devised that computes the degree of matching between the output
given by a FSA and the known optimal solution. This measure takes into
account the particulars of relevance, irrelevance, redundancy and size
of synthetic data sets.

Our results illustrate the pitfall in relying in a single algorithm
and sample data set, very specially when there is poor knowledge
available about the structure of the solution or the sample data size
is limited. The results also illustrate the strong dependence on the
particular conditions in the data set description, namely the amount
of irrelevance and redundancy relative to the total number of
features. Finally, we have shown by a simple example how the
evaluation of a feature subset can be misleading even when using a
reliable inducer. All this points in the direction of using
\textit{hybrid} algorithms (or principled combinations of algorithms)
as well as resampling for a more reliable assessment of feature subset
performance.

This work can be extended in many ways, to carry up more general
evaluations (considering richer forms of redundancy) and using other
kinds of data (e.g., continuous data). A specific line of research is
the corresponding extension of the scoring criterion.

\bibliographystyle{apalike}

\section*{Appendix}
{\bf Proposition.}  The score fulfills the two conditions: 
\begin{itemize}
\item [a)] $S_X({\cal A}) = 0 \Longleftrightarrow {\cal A} = X_I$
\vspace*{-0.25truecm}
\item [b)] $S_X({\cal A}) = 1 \Longleftrightarrow {\cal A} = X^*$
\end{itemize}

{\footnotesize
\noindent {\em Proof}: 

\noindent
a) \fbox{$\Longleftarrow$} Let ${\cal A} =
X_I$. Then $S_X({\cal A}) = S_X(X_I)$; since ${\cal A}_I = X_I
\cap {\cal A} = X_I \cap X_I = X_I$, we have $I=0$; since ${\cal A}_R
= {\cal A}_{R'}$ we have $R = R' = 0$. Thus $S_X(X_I)=0$.

\fbox{$\Longrightarrow$} Suppose $S_X({\cal A}) = 0$; since all terms
that make up $S_X({\cal A})$ are non-negative, it is necessary that
all of them are zero. Now $R=0$ implies ${\cal A}_R \cup {\cal A}_{R'}
= \emptyset$, which implies ${\cal A}_R = {\cal A}_{R'} = \emptyset$;
then ${\cal A} = {\cal A}_I$; thus ${\cal A} = {\cal A} \cap X_I$ and
hence ${\cal A} \subseteq X_I$. Since $I$ must be zero, $|{\cal
  A}_I| = |X_I|$ and therefore ${\cal A} = X_I$.

\noindent
b) Suppose ${\cal A} = X^*$; then it can be checked that $R=R'=I=1$
and thus $\alpha_R R_{\cal A} + \alpha_{R'} R'_{\cal A} + \alpha_I I_{\cal A}
= \alpha_R + \alpha_{R'} + \alpha_I = 1$. Now this is the only way to
achieve this value, since any other situation ${\cal A} \neq X^*$
leads to either $R,R'$ or $I$ to be less than 1.
}

\end{document}